# Hierarchical Deep Learning Classification of Unstructured Pathology Reports to Automate ICD-O Morphology Grading

Waheeda Saib
IBM Research Africa
Johannesburg, South Africa
WSaib@za.ibm.com

Tapiwa Chiwewe
IBM Research Africa
Johannesburg, South Africa
tchiwewe@za.ibm.com

Elvira Singh
National Cancer Registry
Johannesburg, South Africa
elvira.singh@nion.nhls.ac.za

## Abstract

Timely cancer reporting data are required in order to understand the impact of cancer, inform public health resource planning and implement cancer policy especially in Sub Saharan Africa where the reporting lag is behind world averages. Unstructured pathology reports, which contain tumor specific data, are the main source of information collected by cancer registries. Due to manual processing and labelling of pathology reports using the International Classification of Disease for oncology (ICD-O) codes, by human coders employed by cancer registries, has led to a considerable lag in cancer reporting. We present a hierarchical deep learning classification method that employs convolutional neural network models to automate the classification of 1813 anonymized breast cancer pathology reports with applicable ICD-O morphology codes across 9 classes. We demonstrate that the hierarchical deep learning classification method improves on performance in comparison to a flat multiclass CNN model for ICD-O morphology classification of the same reports.

## 1   Introduction

Cancer is one of the leading causes of death worldwide, outstripping malaria and HIV combined. According to the World Health Organization (WHO), the global cancer deaths in 2018 are estimated at 9.6 million and approximately 70% of these deaths occur in low and middle income countries [1]. In Sub-Saharan Africa, cancer is emerging as a public health challenge with the cancer burden estimated to increase by more than 80% by 2030 [2]. Lack of awareness of the cancer burden, a shortage of medical oncologists, and limited access to diagnostic services [3] contribute to the low survival rate of cancer patients in Sub Saharan Africa, in comparison to high income countries, as patients are diagnosed and treated with advanced stages of the disease [4]. Diagnostic services play a vital role in the treatment and management of oncology patients. Accurate tumour diagnosis, including the assessment of tumour margins, biomarker identification, primary tumour site and tumour morphology are part of the pathologist's role. Pathology reports are a foundational source of information for cancer registries, which play a significant role in reporting nationwide cancer statistics and raising global awareness of the impact of cancer. Cancer registries report on diagnosed cancer cases and incidence rates, which are significant for public health planning in resource constrained areas. Due to the manual process of coding pathology reports into appropriate International Classification of Disease for Oncology (ICD-O) codes and the magnitude of reports received annually, a considerable lag time exists in national cancer statistics reporting. Hence there is an evident need to develop methods that automate the manual processes employed by cancer registries, to classify pathology reports into relevant ICD-O codes.

## 2   Related Work

Several automated approaches have been proposed for labelling cancer reports. In [7], a literature review is provided of rule-based information retrieval (IR) and natural language processing (NLP) techniques. Rule-based systems were not generalizable across cancer domains and struggled with the variability in structure across reports. While the authors suggest machine learning (ML) approaches may be necessary for "complex and variable types of information", they conclude by voicing a "strong preference for rule-based systems over 'black box' ML models in clinical practice." As the field of ML and deep learning (DL) applied to natural language processing (NLP) advances, there is increased explainability of black box models that outperform rule-based solutions. A recent publication [9] demonstrates that convolutional neural networks (CNNs), using word embeddings, consistently perform better than classical ML approaches using term frequency-inverse document frequency (TF-IDF) for IR and classification of pathology reports by primary tumour site. Typically, classification by tumour site is done using ICD-O codes, which are two part codes, including a topographical code (anatomical site of origin of the tumour), and a morphological code (cell type of tumour or histology and behaviour, i.e., malignant or benign). ICD-O classification is hierarchical in nature, with the top layer representing disease categories and the sub layers representing narrowing disease paths.

We thus use this knowledge as a basis to explore hierarchical CNNs for ICD-O classification of cancer reports. Hierarchical classification methods with conventional machine learning models such as SVM and Naıve Bayes, were used to create hierarchical tree structures of expert binary classifiers that solve a multiclass problem in [10]. Another approach explores a hierarchy of sub level multiclass classifiers, to attain better performance on multiclass problems [11]. Work on hierarchical text classification in deep learning, explores the different combinations of specialized deep neural networks, recurrent neural networks and CNNs as the first and second layer models for document classification [12]. With respect to hierarchical classification in the medical domain, [13] employs a 'hierarchical mixture of experts' approach that uses neural networks as the parent classifier and sub level linear classifiers such as Widrow-Hoff and Exponential Gradient to classify Medline medical abstracts.

To the best of our knowledge, the present work is the first to apply a hierarchical deep learning classification method using multiclass and binary CNN models to classify breast cancer pathology reports by ICD-O morphology codes. Our hypothesis is that for NLP tasks such as ICD-O code classification which are hierarchical in nature, employing a hierarchy of the best performing deep learning models trained for classification of free text pathology reports into ICD-O morphology codes, will achieve greater accuracy for an "unseen" dataset than a flat multiclass model. This could potentially pave the way for large scale deployment in a national cancer registry, where pathology reports cover a wide range of cancer types

## 3      Materials and Methods

### 3.1      Data description and Preparation

The cancer pathology dataset used in our analysis, consists of 1813 anonymized, unstructured breast cancer pathology reports. Each report is manually assesesed and tagged by trained human coders with one of several morphgical codes. The morphogy code structure comprises five digits, the first four represent the cancer cell type and the last digit after the slash shows the behaviour of the tumour [14]. These tags were used as labels for our classification models. There are 48 ICD-O morphological classes present in the dataset, of which we select the nine classes that contain reports in the range of 12 to 111 reports with the exception of class 8500/3 containing 1417 reports. The cancer pathology report dataset was formatted in extensible mark-up language (XML), which is a convention used to store data in a semi-structured machine and human readable form. In our dataset, the XML tags demarcated the report text and type. The first step in the data preparation process extracted the report text from the XML tags. Thereafter reports are then sent to an anonymization program to verify that the data has been appropriately de-identified. The Afrikaans only reports, numbers, stop words and punctuation symbols are removed from the report text which is then converted to lowercase and tokenized. We extract the top 1400 TF-IDF features, which were determined during hyper parameter testing, from the pathology reports. The TF-IDF features are used to filter the report text of the pathology reports, effectively reducing the document length while retaining the most significant words.

### 3.2      Experimental Design

The labelling of the breast cancer pathology reports into 9 ICD-O morphology codes is treated as a multiclass classification task. Due to the nature of the breast cancer pathology reports having highly correlated terms and the number of ICD-O categories that may be assigned to each report, makes this a difficult problem to solve. To reduce the complexity of the multiclass problem, we deconstruct it into an unconventional hierarchical classification model, that is structured as a top down convolutional neural network ensemble of binary and multiclass classifiers. We establish the baseline for model comparison as the flat nine class ICD-O multiclass CNN. In order to identify the classes that should be designated to the second level of classifiers, we use the distribution of the morphology codes in the dataset and the confusion matrix of the flat 9 class, multiclass CNN.

### 3.2.1 CNN for Text Classification

Convolutional neural networks have recently become the standard to benchmark text classification tasks [17]. In NLP tasks, the input is a document matrix, where each row corresponds to a word density vector. CNNs are made up of convolutional and pooling layers that act as a key feature generation mechanism. Convolutions are performed by applying a learned filter $w$ to a receptive field, described in terms of a sliding window of $h$ words. By applying the learned filter on every instance, on the $h$ word sliding window of the document results in feature maps that capture relevant properties of the words within each window. The feature maps are subsampled by taking the maximum values per dimension over different window results, in the subsequent pooling layer [16]. This approach encourages location invariance and decreases dimensionality of the output as it is passed to subsequent layers of the network. Through this mechanism, we learn key words and sequence phrases that contribute to the different categories in the ICD-O classification task.

### 3.2.2 Hierarchical Deep Learning

Our contribution in this paper is a novel hierarchical deep learning classification method to automate the labelling of free text cancer pathology reports with ICD-O morphology codes. We propose an approach to deconstruct the multiclass classification problem into a top down tree structure of specialized CNN classifiers that improve on the performance of a traditional CNN classifier on the selected ICD-O morphology classes. In the first level of the hierarchical method, illustrated in Figure. 1, the parent classifier is a multiclass CNN that is designed to take as input all 9 ICD-O report classes and predict the group the report belongs to. The second level child classifiers consist of a multiclass and binary CNN model. A base CNN architecture for both multiclass and binary model is used, that randomly initializes an embedding layer with the 1400 TF-IDF word features before applying the architecture described in [17]. In this architecture, the convolutional layer computes 100 feature maps for each window of size 3, 4 and 5 words. On the resulting feature map maxpooling is applied, and a subsequent dropout rate of 0.5 is applied to this result. The result is then fed into a hidden layer before applying softmax classification. The model was trained for 147 epochs, 75 batch size using the adadelta optimizer. . The way in which the hierarchical ensemble is constructed is presented in Figure. 1

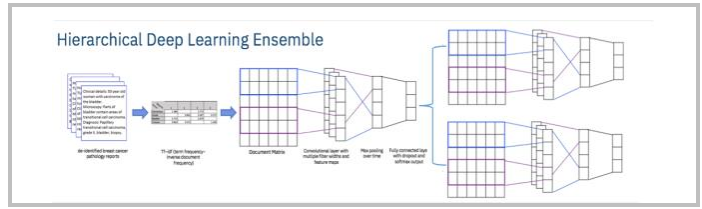

Figure 1: Hierarchical CNN Ensemble

### 3.2.3 Hyper parameter optimization and Evaluation

For the CNN models, initial experiments on the architectures were based on parameters and baselines established in, the sensitivity analysis of CNNs [18]. The dataset split employed by the model was 80/10/10 for train, validation and test set. We implemented 10-fold cross validation on the CNN model variations for different class subsets. During each fold, the model weights that performed well on the validation set were saved. Performance was measured using the F1-micro and F1-macro metric. The held-out test set during cross validation was used to evaluate the final model. The results were analyzed to understand the effect ICD-O classes had on overall performance. The model results from each fold were merged to perform bootstrapping with replacement to compute 95% confidence intervals as presented in Table 1. For the hierarchical CNN, the selection of the best models for each classifier in this ensemble, was determined during cross validation by performance on the validation set. An automated pipeline was developed to evaluate the hierarchical ensemble's performance against the flat multiclass model on the unseen test set.

# 4 Results and Discussion

To evaluate the performance of the flat 9 class multiclass CNN, we analyze its confusion matrix and note that due to the unbalanced distribution of classes, the model has specialized in classifying the majority class, 8500/3, which is the most commonly labelled class "infiltrating ductal carcinoma". To address this imbalance in the dataset, we remove the class 8500/3 to assess performance of the multiclass CNN on the remaining 8 morphology classes and perform binary one vs all (OVA) to specialize classification of the majority class 8500/3 using a binary CNN.

We analyze the performance of the binary and multiclass classifiers for the proposed class breakdown to establish the models to use as the child classifiers of the hierarchical classification method. Examining the mean F1 macro and micro scores from cross validation shows that the 8-class multiclass CNN based on our modified CNN architecture is the best model when compared to a traditional CNN architecture with the F1 micro of 0.791 and an F1 macro score of 0.623. Hence this is chosen as one our specialized child classifiers. With the majority class, a binary OVA is explored and optimized to attain results of F1 micro at 0.892 and F1 macro at 0. 838. This is then used as the second specialized child classifier. For the parent classifier, the 9 classes are partitioned into two groups and applied to the CNN architecture with a final softmax layer.

Table 1: Comparison of Flat Multiclass and Hierarchical CNN Ensemble

| ICD-O Classes | Classifier | F1 Micro | F1 Macro |
|---|---|---|---|
| 9 | Multiclass CNN | 0.887 (0.872, 0.902) | 0.565 (0.529, 0.601) |
| 8 | Multiclass CNN Child Classifier | **0.791** (0.753, 0.827) | **0.623** (0.578, 0.668) |
| 8 | Traditional Multiclass CNN | 0.773 (0.732, 0.811) | 0.609 (0.564, 0.659) |
| 2 | Binary OVA CNN Child Classifier | **0.892** (0.877, 0.906) | **0.838** (0.816, 0.858) |
| Best Model Performance | | | |
| 9 | Multiclass CNN | 0.896 | 0.581 |
| 9 | Hierarchical CNN | **0.918** | **0.692** |

The results of the hierarchical CNN classification and flat multiclass model are in Table 1. The performance of the hierarchical CNN classification method achieves an F1 micro score of 0.918 and an F1 macro score of 0.692 in comparison to the 9-class multiclass CNN which achieved an F1 micro score of 0.896 and F1 macro score of 0.581. This demonstrates that by decomposing the multiclass problem into specialized tree structure of deep learning models, optimized to perform well on the relevant class subsets, results in a performance increase.

# 5 Conclusion

In this paper, we propose an automated method to achieve increased performance on a multiclass text classification problem. We deconstruct a multiclass classification problem into a tree like structure of specialized classifiers optimized to perform well on different class partitions. We demonstrate a way to determine the components of the hierarchical ensemble by analysing the class distribution and observing the model performance via the confusion matrix. This work diverges from the initial methods of hierarchical classification that decompose a multiclass problem into a tree of binary models. Initial results show that our presented hierarchical approach improves on a flat multi-class approach. Further exploration of this hierarchical model of classification will be employed across the entire ICD-O coding structure, which itself is hierarchical in nature. These results will enhance the classification of pathology reports by cancer registries, resulting in faster access to these vital data, that may contain trends that call for reallocation of scarce resources.